\documentclass[smallabstract,smallcaptions]{dccpaper}
\usepackage{fancyhdr}
\usepackage{epsfig}
\usepackage{citesort}
\usepackage{amsmath}
\usepackage{amssymb}
\usepackage{color}
\usepackage{url}
\usepackage{amsmath,amsfonts}
\usepackage{algorithmic}
\usepackage{algorithm}
\usepackage{array}
\usepackage{subfig}
\usepackage{paralist}
\usepackage{url}
\usepackage{textcomp}
\usepackage{stfloats}
\usepackage{verbatim}
\usepackage{graphicx}
\usepackage{cite}
\usepackage{booktabs}
\usepackage{hyperref}
\def \NOTE [#1]{\textcolor{blue}{(\textit{#1})}}
\usepackage{pifont}
\usepackage{bbding}
\usepackage{float}
\usepackage{caption}
\usepackage{graphicx, subfig}
\usepackage{amsthm,amsmath,amssymb}
\usepackage{mathrsfs}
\usepackage{multirow}

\newlength{\figurewidth}
\newlength{\smallfigurewidth}

\begin{document}

\title
{\large
\textbf{Extreme Image Compression Using Fine-tuned VQGANs}
}

\author{%
Qi Mao$^{\ast}$, Tinghan Yang$^{\ast}$, Yinuo Zhang$^{\ast}$, Zijian Wang$^{\ast}$, \\
Meng Wang$^{\dag}$, Shiqi Wang$^{\dag}$, Libiao Jin$^{\ast}$ and Siwei Ma$^{\ddagger}$\\
{\small\begin{minipage}{\linewidth}\begin{center}
\begin{tabular}{ccc}
$^{\ast}$Communication University of China & $^{\dag}$City University of Hong Kong & $^{\ddagger}$Peking University
\end{tabular}
\end{center}\end{minipage}}
}

\maketitle
\thispagestyle{empty}

\begin{abstract}
Recent advances in generative compression methods have demonstrated remarkable progress in enhancing the perceptual quality of compressed data, especially in scenarios with low bitrates.
However, their efficacy and applicability to achieve extreme compression ratios ($<0.05$ bpp) remain constrained.
In this work, we propose a simple yet effective coding framework by introducing vector quantization (VQ)--based generative models into the image compression domain.
The main insight is that the codebook learned by the VQGAN model yields a strong expressive capacity, facilitating efficient compression of continuous information in the latent space while maintaining reconstruction quality.
Specifically, an image can be represented as VQ-indices by finding the nearest codeword, which can be encoded using lossless compression methods into bitstreams. 
 We propose clustering a pre-trained large-scale codebook into smaller codebooks through the K-means algorithm, yielding variable bitrates and different levels of reconstruction quality within the coding framework.
Furthermore, we introduce a transformer to predict lost indices and restore images in unstable environments.
Extensive qualitative and quantitative experiments on various benchmark datasets demonstrate that the proposed framework outperforms state-of-the-art codecs in terms of perceptual quality-oriented metrics and human perception at extremely low bitrates ($\le 0.04$ bpp).
Remarkably, even with the loss of up to $20\%$ of indices, the images can be effectively restored with minimal perceptual loss.
\end{abstract}

\section{Introduction}
With the ever-increasing amount of visual data being generated at an unprecedented pace, the demand for highly efficient and effective compression algorithms has become increasingly crucial.
However, under minimal network bandwidth, the signal-oriented traditional image/video compression codecs (\textit{e.g.}, BPG~\cite{BPG}, and the latest video coding standard VVC \cite{VVC}) inevitably adopt large scalar quantization steps, resulting in a significant loss of texture information with unacceptable blurring and blocking artifacts.
%

To bridge the gap of shallow bitrate scenarios, recent image coding methods~\cite{rippel2017real,santurkar2018generative,agustsson2019generative,lee2020training,mentzer2020high,iwai2021fidelity,chang2019layered,chang2021thousand,chang2022conceptual,chang2022consistency} leverage the power of generative models~\cite{goodfellow2020generative,kingma2013auto} to reconstruct the human-favored decoded image/video.
There are currently two main perspectives in such generative compression: the first~\cite{rippel2017real,agustsson2019generative,mentzer2020high,iwai2021fidelity} involves using a conditional GAN~\cite{mirza2014conditional} as an additional distortion term to optimize deep learning-based end-to-end neural codecs. 
This category of methods enhances the reconstruction of texture details in the decoded image through adversarial training. However, their effectiveness in achieving high compression ratios, especially below $0.05$ bpp, remains limited.
Another line~\cite{santurkar2018generative,chang2019layered,chang2021thousand,chang2022conceptual,chang2022consistency,changIJCV} aims to compress images into compact feature representations at the encoding end and generate decoded images with the aid of GANs, achieving visual pleasing reconstruction even at extremely low compression ratios.
However, without additional training, such approaches~\cite{chang2019layered, chang2021thousand,chang2022conceptual,chang2022consistency,changIJCV} have difficulty reconstructing the original image with large semantic information gaps against the training dataset.
For instance, codecs optimized for face images may not perform well on natural scenario images. 
As such, their practical use in extremely low-bitrate scenarios is hindered by poor generalization ability.
Recently, vector quantization (VQ)--based generative models~\cite{esser2021taming,chang2022maskgit}, which utilize discrete image representations, have been well used in image generation tasks.
Despite their success in other generation fields, VQ-based generative models have received relatively less attention in the image compression domain.
In this work, we reveal a surprising finding:
The learned codebook of the VQGAN model~\cite{esser2021taming}, which has been trained on a large-scale dataset, exhibits a powerful and robust representational capacity.
Accordingly, we propose a simple yet effective coding framework by directly integrating VQ-indices compression into the VQGAN model, which yields a significant advancement in the capability of extreme image compression and generalizability across various semantics and resolutions of images.
In particular, the VQ-indices map is obtained by identifying the nearest sample in the learned codebook, which is then encoded into a bitstream using arithmetic compression. 
To enable variable bitrates, we propose clustering a pre-trained large-scale codebook using the K-means algorithm, resulting in a series of smaller codebooks. 
%
%
As such, the image can be represented by various VQ-indices maps, enabling variable bitrates and different levels of reconstruction quality.
Moreover, given the potential impact of bitstream loss on the decoding process within an unstable transmission environment, we propose to leverage the second-stage transformer of the VQGAN model to predict missing indices.
It can be effectively predicted based on context indices that adhere to the underlying discrete distribution, thereby effectively circumventing image reconstruction failure due to the loss of bitstreams.
The main contributions in this paper can be summarized as follows:
\begin{compactitem}
\item We present a simple yet effective coding framework that utilizes the VQGAN model in developing a novel extreme image compression framework.

\item We propose a K-means clustering approach to compress the large-scale codebook into a smaller new codebook, which enables variable bitrates and levels of reconstruction quality within our framework.
Furthermore, a second-stage transformer is leveraged to predict missing indices, enhancing the resilience of the proposed framework in unstable transmission.

\item Both qualitative and quantitative results demonstrate that the proposed framework achieves a significant improvement compared to state-of-the-art codecs about both perceptual quality metrics and human-viewed under extremely low bitrates ($\le 0.04$bpp).
Notably, even in cases where up to $20\%$ of indices are lost, the images can be efficiently restored with minimal perceptual loss, thanks to the collaboration with the generative transformer model.
\end{compactitem}

\begin{figure*}[!t]
    \centering
    \includegraphics[width=0.98\linewidth]{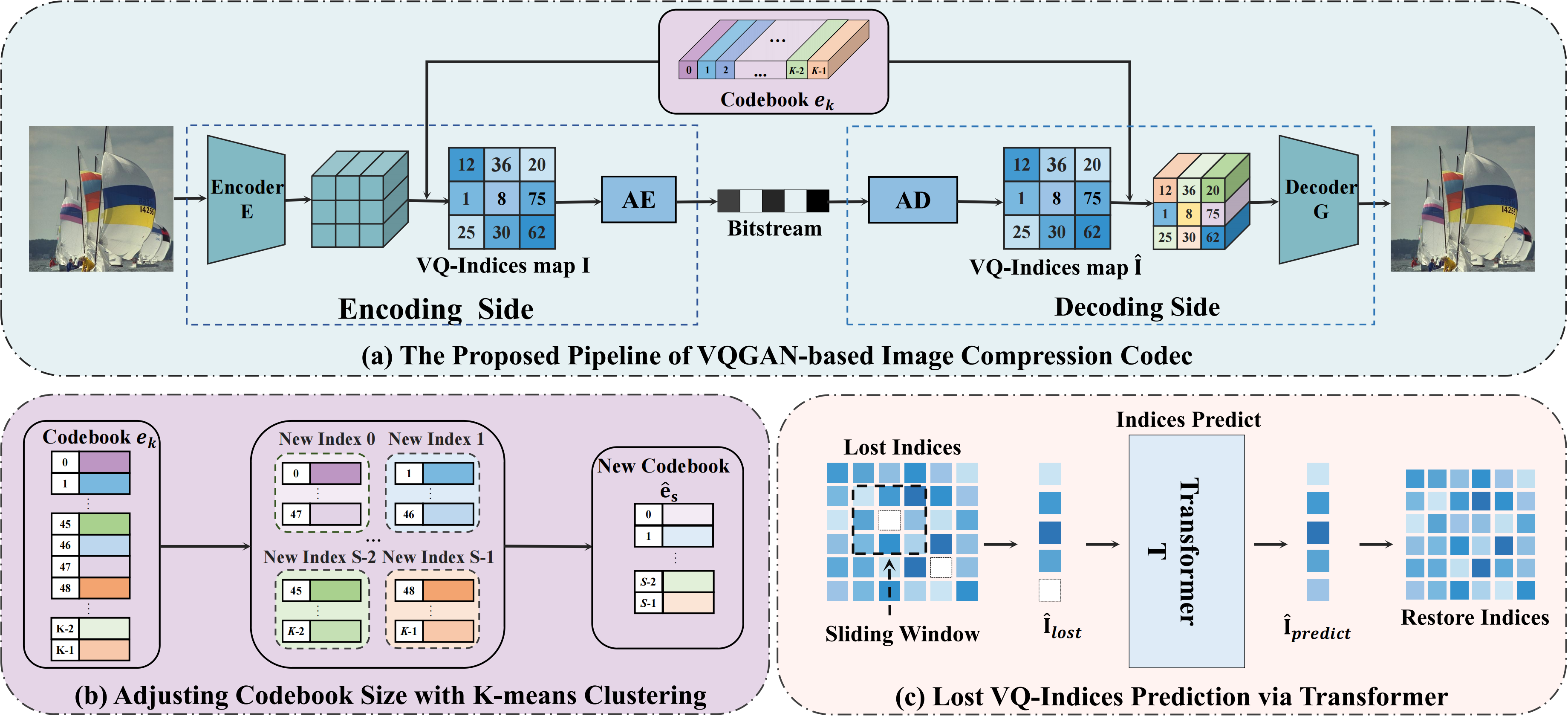}    \caption{Overview of the proposed VQGAN-based image coding framework.
    }
    \label{fig:framework}
\end{figure*}

\section{Proposed VQGAN-based Extreme Image Codec}

In this work, we aim to compress images at ultra-low bit rates while maintaining the high-perceived quality of the reconstructed images. 
Our framework is built upon the discrete representations and the generative capacity of VQGAN models.
As illustrated in Fig.~\ref{fig:framework}, the entire framework consists of four key components:
\begin{compactitem}
\item The encoder $E$: extract the input image $\mathbf{x}\in \mathbb{R}^{H\times W\times 3}$ into a latent representation $\mathbf{z}\in \mathbb{R}^{\frac{H}{M}\times\frac{W}{M}\times n_z}$.
\item The codebook $\mathbf{e}_k\in \mathbb{R}^{n_z},k\in 1,2,\cdots, K$: map the latent representation $\mathbf{z}$ into a sequence of VQ-indices and invert it back to quantized latent representation $\mathbf{z}_q\in \mathbb{R}^{\frac{H}{M}\times\frac{W}{M}\times n_z}$ through the nearest neighbor lookup.
The proposed K-means clustering algorithm method compresses the large-scale codebook into a smaller one, enabling variable bitrates and varying reconstruction quality.
\item The decoder $G$: synthesize the quantized latent representation $\mathbf{z}_q$ into a reconstructed image $\mathbf{\hat{x}}\in \mathbb{R}^{H\times W\times 3}$.
\item The transformer $T$: predict the missing index based on the context indices.
\end{compactitem}
On the encoding side, VQ-indices are compressed into the final bitstream using lossless compression techniques. 
On the decoding side, we decode the index sequences from the bitstream and convert the decoded VQ indices back into the quantized latent representation $\mathbf{z}_q$ by searching the codeword from the codebook.
We detail the key components of the proposed framework below.

\subsection{VQ-indices Compression}
\label{sec:vq-index-compression}
Unlike existing image compression methods~\cite{cheng2020learned,theis2017lossy} that adopt scalar-quantization, we leverage the power of VQ to construct a codebook of representative vectors for latent representations $\mathbf{z}$.
In particular, the codebook $\mathbf{e}_k$ is then used to encode the latent representation by replacing each position of a vector with the \textbf{index} of the closest representative vector by Euclidean distance, which results in a highly compressed version of the latent representation with minimal loss of quality:
\begin{equation}\label{eq:1}
\mathbf{I}_{i j}= \underset{k\in 1,2,\cdots, K}{\operatorname{argmin}}\left\|\mathbf{z}_{i j}-\mathbf{e}_k\right\|^2,
\end{equation}
where $i$ and $j$ denote the position of vector $\mathbf{z}_{i j}$ in latent representation $\mathbf{z}$,  $\mathbf{I}_{i j}$ represents its corresponding index, and $K$ indicates the size of codebook.
As demonstrated in Fig.~\ref{fig:framework}(a), \textbf{an input image $\mathbf{x}$ can be efficiently represented by VQ-indices map $\mathbf{I}\in \mathbb{R}^{\frac{H}{M}\times\frac{W}{M}}$, which significantly reduces the data amount.}
Then, we adopt the widely used arithmetic coding to compress VQ-indices into bitstreams, further reducing the data size.
After decoding the compressed bitstream, the reconstructed latent vectors $\mathbf{z}_q$ are generated by searching for their corresponding code words based on their indices.
Finally, the reconstructed image $\hat{\mathbf{x}}$ is synthesized by the decoder $G$.

\subsection{Adjusting Variable Bitrates }
The quality of the codebook in our proposed framework plays a crucial role in determining the compression performance of the entire system. 
As such, we develop a rate control strategy that
utilizes the K-means algorithm to \textbf{cluster the pre-trained large-scale trained codebook into the smaller one},
thereby enabling the flexible adjustment of compression bitrates by altering the size of the codebook.
%
%
%
%
Subsequently, the newly generated codebook $\hat{\mathbf{e}}_{s}$ derived from the K-means clustering algorithm can serve as a starting point for further fine-tuning, enabling faster convergence during subsequent optimization while ensuring codebook quality.
Finally, we can obtain new codebooks of varying sizes, each with a corresponding set of VQ-indices to represent the compressed image, as illustrated in Fig.~\ref{fig:framework}(b).
%

\begin{figure*}[!t]
\centering
\includegraphics[width=0.98\linewidth]{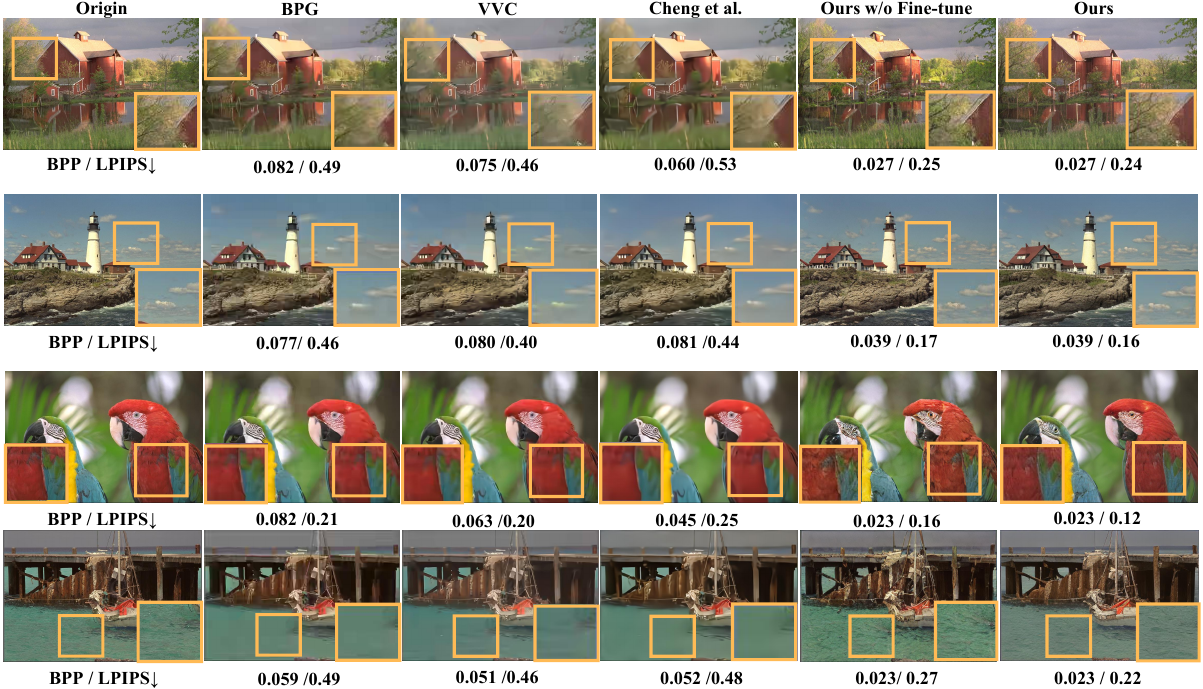}
\\
\caption{The qualitative comparison results of BPG, 
 VVC, Cheng \textit{et al.}, ours w/o fine-tuning and our method on the Kodak dataset. 
 %
 }
 \label{fig:VQ-GANs}
\end{figure*}

\subsection{Lost VQ-indices Prediction}
\label{sec:index-lost}
In an unreliable transmission environment, lost packets may result in the loss of indices, which can cause incorrect decoding of bitstreams. 
However, our proposed image codec framework, which utilizes the VQ-indices transformer, can \textbf{accurately predict lost indices at the decoder end}. 
This approach enhances the robustness of the codec in dealing with unreliable network transmission and ensures that the decoded bitstreams remain accurate.
We flatten the VQ-indices map $\mathbf{\hat{I}}$, and denote it as $[\mathbf{\hat{I}}_i]^{N}_{i=1}$, where $N$ indicates the total length. 
Subsequently, the VQGAN's second-stage transformer is trained to predict the probability distribution of the next possible indices $p(\mathbf{\hat{I}}_i|\mathbf{\hat{I}}_{<i})$. 
The objective is to maximize the log-likelihood of the data representation, which can be expressed as follows:
\begin{equation}\label{eq:9}
\mathcal{L}_{T}=\mathbb{E}_{\mathbf{x}\sim p(\mathbf{x})}{\big[} -\log p(\mathbf{\hat{I}}){\big]},
\end{equation}
where $p(\mathbf{\hat{I}})=\prod_{i}p(\mathbf{\hat{I}}_i|\mathbf{\hat{I}}_{<i})$.
To simulate the potential loss of indices during transmission, we incorporate a \textbf{masking} procedure.
Specifically, we apply a binary mask $M=[m_i]^{N}_{i=1}$ as follows:
For $m_i = 1$, the corresponding index $\mathbf{\hat{I}}_i$ is replaced by a special $[mask]$ token to indicate that it has been lost.
Conversely, if $m_i = 0$, then $\mathbf{\hat{I}}_i$ is left unchanged.
The mask process is controlled by a mask ratio ($\alpha \in [0,1]$), which determines the number of missing indices as $\alpha \cdot N$, denoted as $\mathbf{\hat{I}}_{lost}$.
During the restoration stage, as depicted in Fig.~\ref{fig:framework}(c), a sliding window input strategy is employed to input a $16 \times 16$ window centered on $\mathbf{\hat{I}}_i$ at each prediction. Only the indices before $i$ in the $16\times 16$ window are included as input due to the autoregressive nature of the transformer, which is consistent with the decoding process.
Finally, the restored image $\mathbf{\tilde{x}}$ is generated by feeding the predicted index sequence $\mathbf{\hat{I}}_{predict}$ into the decoder $G$.

\section{Experimental Results}
\subsection{Experimental Settings}
\label{sec:Setup}
\paragraph{DataSets.}
%
The proposed framework is trained on the ImageNet dataset~\cite{deng2009imagenet}, which contains $1.2$ million images distributed across $1,000$ distinct categories.
To assess the performance of the proposed model, we evaluate two commonly used datasets in image compression: the Kodak dataset~\cite{kodak}, which includes $24$ natural uncompressed $512\times768$ or $768\times512$ images; and the CLIC2020 dataset~\cite{CLIC}, comprising $250$ images that exhibit varying lighting conditions and dynamic range, with resolutions ranging from $320\times240$ to $4032\times3024$.

\paragraph{Implementation Details.}
%
%
%
We adopt $M=16$ to downsample the image $\mathbf{x}$ into the latent representation $\mathbf{z}$. 
First, the weights of encoder $E$, decoder $G$, and the discriminator $D$ are initialized by the officially provided pre-trained VQ-GAN model~\footnote{{https://github.com/CompVis/taming-transformers}}. 
To enable the proper bitrate range, we perform K-means clustering on the size of $16384$ codebook of the pre-trained model, reducing the new codebooks into the size, ranging from $\{2048, 1024, 512, 256, 128, 64, 32, 16, 8\}$.
We then fine-tune the entire framework using the default settings and training losses as \cite{esser2021taming}, which takes $13$ hours on two NVIDIA Tesla-A100 GPUs.

\begin{figure*}[!t]
    \centering
    \includegraphics[width=0.98\linewidth]{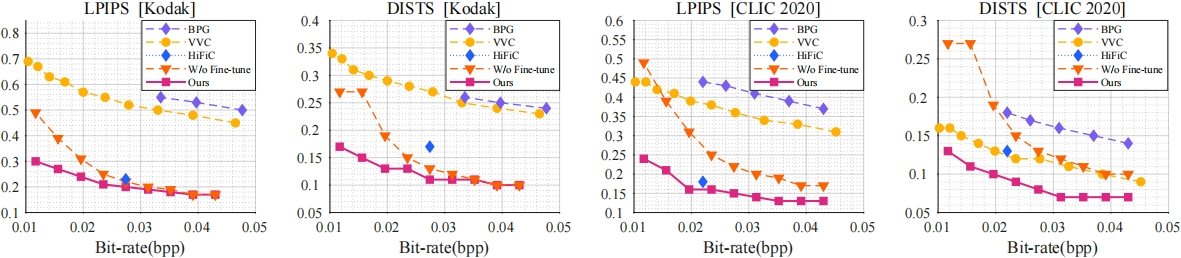}
   \caption{The R-D performance of BPG, VVC, Cheng \textit{et al.}, FCC, HiFiC, ours w/o fine-tuning and the proposed method on the Kodak dataset, and the CLIC2020 dataset.
   }
\label{fig:RD}
\end{figure*}
\begin{table*}[!t]
\vspace{-3mm}
\begin{center}
\caption{BD-rate and BD-metric relative to the
BPG, VVC, Cheng~\textit{et al.}, and ours w/o fine-tuning respectively, where LPIPS is used
as the distortion metric in BD-metric.} \label{tab:tab1}
\begin{tabular}{c|cc|cc}
\hline
\multirow{2}{*}{Baseline} & \multicolumn{2}{c|}{Kodak}               & \multicolumn{2}{c}{CLIC}                 \\ \cline{2-5} 
                          & \multicolumn{1}{c|}{BD-rate}  & BD-LPIPS & \multicolumn{1}{c|}{BD-rate}  & BD-LPIPS \\ \hline
VVC                       & \multicolumn{1}{c|}{-98.55\%} & -0.33    & \multicolumn{1}{c|}{-93.02\%} & -0.21    \\ \hline
BPG                       & \multicolumn{1}{c|}{-99.99\%} & -0.36    & \multicolumn{1}{c|}{-99.74\%} & -0.27    \\ \hline
HiFiC                     & \multicolumn{1}{c|}{-11.62\%} & -0.02    & \multicolumn{1}{c|}{-8.23\%}  & -0.01    \\ \hline
Cheng et. al              & \multicolumn{1}{c|}{-97.93\%} & -0.37    & \multicolumn{1}{c|}{-98.90\%} & -0.25    \\ \hline
Ours W/o Fine-tune        & \multicolumn{1}{c|}{-21.57\%} & -0.06    & \multicolumn{1}{c|}{-25.59\%} & -0.06    \\ \hline
\end{tabular}
\end{center}
\end{table*}

\paragraph{Evaluation Metrics.}
Traditional objective quality assessment methods, such as PSNR and SSIM, have been primarily devised for the calculation of pixel-level distortions. However, these methods may not be suitable for assessing perceptual quality.
Therefore, we integrate recent approaches \textit{i.e.} the learned perceptual image patch similarity (LPIPS)~\cite{zhang2018unreasonable} and the deep image structure
and texture similarity (DISTS)~\cite{ding2020image} to assess perceptual quality, which better aligns with the way humans perceive images.
All of the above metrics are the smaller, the better.
Furthermore, we utilize bits per pixel (bpp) to evaluate the rate performance. 

\vspace{-3 mm}
\subsection{Compression Performance Comparison}
\paragraph{Compared Methods.}
To evaluate the efficacy of our proposed framework, we compare the proposed method with both
traditional standard and neural-based typical compression frameworks:
%
First, we compare with classic image codec BPG~\cite{BPG} and the latest video coding codec VVC~\cite{VVC}.
%
%
%
%
%
For the typical end-to-end (E2E) codec based on deep learning, we compare with Cheng \textit{et al.} \cite{cheng2020learned} and retrain the model implemented by CompressAI\footnote{https://github.com/InterDigitalInc/CompressAI} to cover
a bitrate range similar to ours.
%
%
Furthermore, we develop an additional baseline that directly adopts the K-means clustering algorithm without any fine-tuning, denoted as ``Ours w/o Fine-tune''.

\begin{figure}[!t]
\centering
\centering
\includegraphics[width=0.98\linewidth]{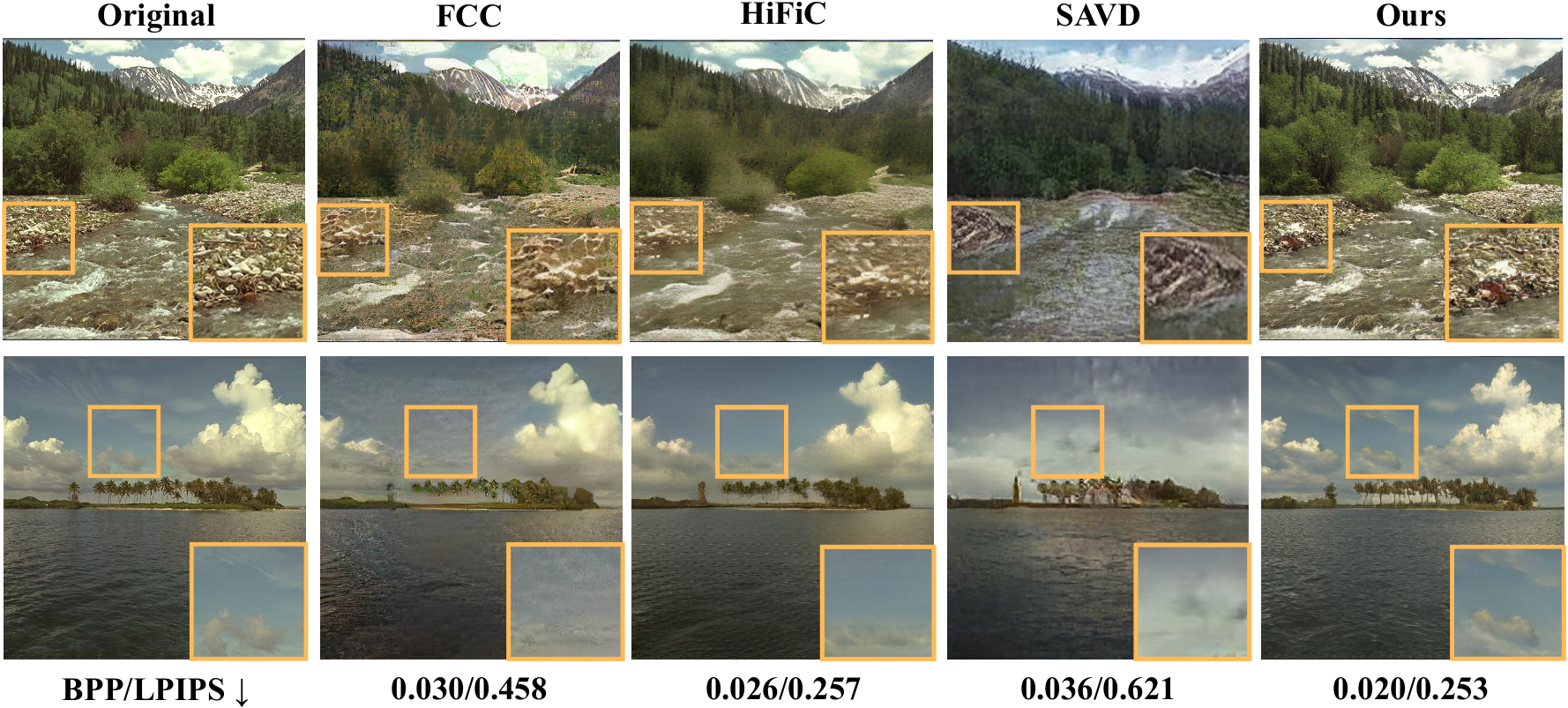}
\caption{Qualitative comparisons of three typical generative compression codecs (\textit{i.e.} FCC, HiFiC, SAVD) and 
the proposed method over the Kodak dataset. 
}
 \label{fig:generative-comparisons}
 \vspace{-5 mm}   
\end{figure}

\paragraph{Qualitative Evaluation.}
Fig.~\ref{fig:VQ-GANs} shows the reconstruction results of various methods on different images on the Kodak dataset~\cite{kodak}, as well as the corresponding bpp and LPIPS.
%
 %
%
It is evident that at extremely low bitrates, VVC~\cite{VVC}, BPG~\cite{BPG}, and Cheng \textit{et al.}~\cite{cheng2020learned} exhibit varying degrees of blurring and missing texture details. 
In contrast, our proposed method reconstructs more texture details, such as trees, ripples, clouds, and feathers, resulting in the lowest LPIPS value. 
Moreover, models obtained using the K-means clustering algorithm can produce relatively rich details even without fine-tuning training, thanks to the inherent expressive power of the pre-trained codebook.
Nevertheless, upon fine-tuning, our proposed method outperforms the former by achieving more color and texture reconstruction consistency. 

\paragraph{Quantitative Evaluation.}
\label{sec: Rate-Distortion Performance}
Our proposed framework is capable of achieving extremely low bitrate compression from $\mathbf{0.01}$ to $\mathbf{0.04}$ bpp while maintaining perceptual quality for a wide range of images with diverse semantics and resolutions across various datasets.
As illustrated in Fig.~\ref{fig:RD}, the experimental results demonstrate the proposed approach exhibits remarkable improvements in perceptual-oriented metrics compared to other compression methods.
To better evaluate the RD performance improvement of our proposed method, we utilize the Bjontegaard metric~\cite{bjontegaard2001calculation}, as demonstrated in Table~\ref{tab:tab1}.
For instance, with the same reconstruction quality of the LPIPS metric, our approach achieves approximately $97.93\%\sim 99.99\%$, $93.02\%\sim99.74\%$ bitrates saving compared to VVC~\cite{VVC}, BPG~\cite{BPG} and Cheng~\textit{et al.}~\cite{cheng2020learned} over the two datasets, respectively. 
These results demonstrate the remarkable advantage of our proposed method achieving high perceptual quality for extremely low bitrate coding.
The performance of models generated by utilizing the K-means clustering algorithm without fine-tuning can yield promising results on the rate-distortion (R-D) curve when the size of the codebook is relatively large. 
Nonetheless, a significant degradation in performance is observed when the codebook size is reduced to below $64$. 
Fine-tuning can effectively enhance the performance of the model under such conditions, with $21.57\%\sim 25.59\%$ bitrates saving, as shown in Table~\ref{tab:tab1}.

\begin{figure}[!t]
\centering
\includegraphics[width=0.99\linewidth]{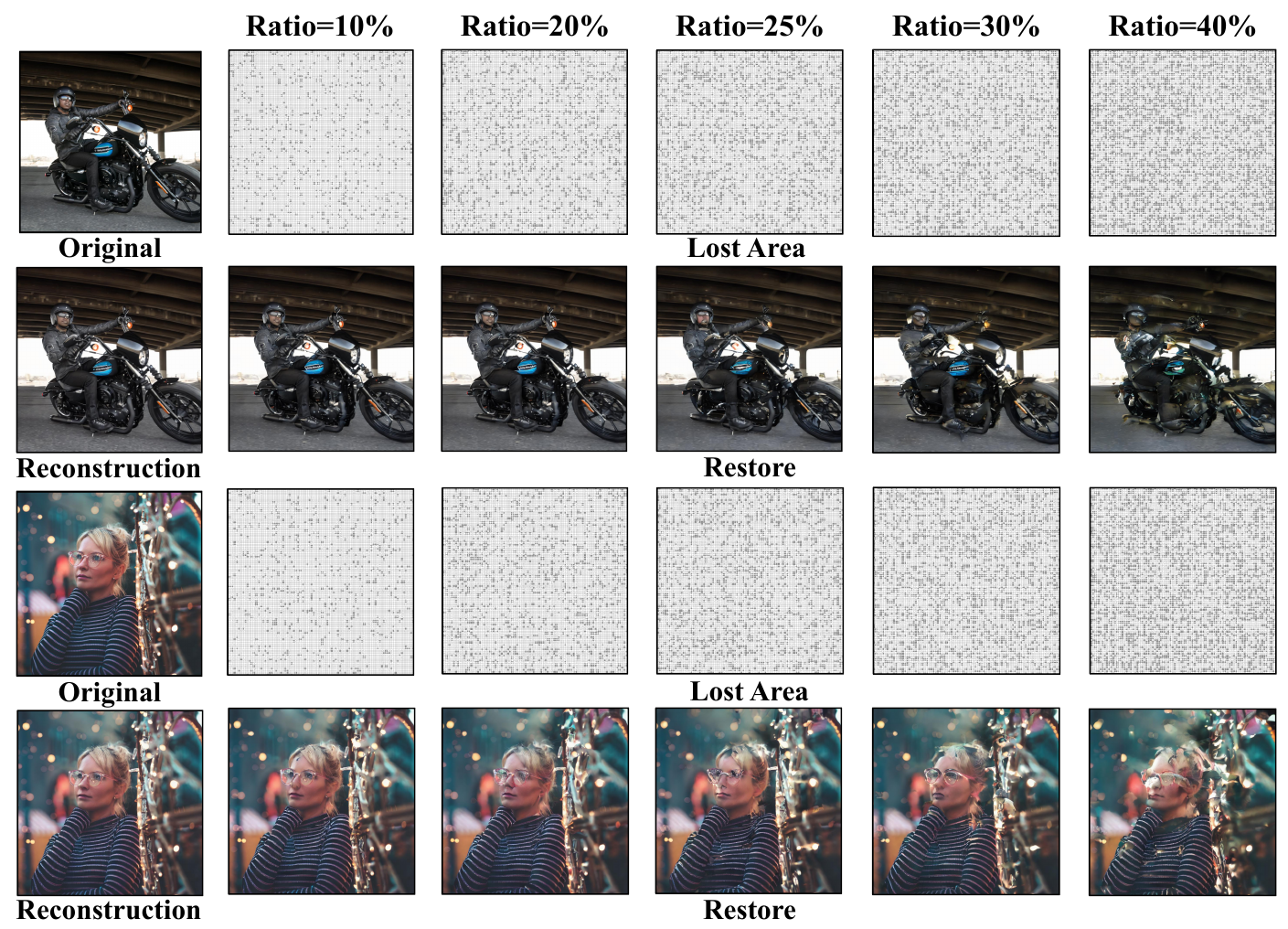}
\\
\caption{Image restoration via lost VQ-indices prediction ranging from $10\%$ to $40\%$.
}
    \label{fig:mask}
     \vspace{-3mm}
\end{figure}

\subsection{Comparisons with Generative Image Compression Codecs.}

To comprehensively assess the efficiency and applicability of the proposed approach vis-a-vis state-of-the-art generative compression codecs, we perform comparative analyses against three representative codecs, as illustrated in Fig.~\ref{fig:RD} and Fig.~\ref{fig:generative-comparisons}. 
FCC~\cite{iwai2021fidelity} and HiFiC~\cite{mentzer2020high} are among the first categorical approaches that utilize adversarial loss to optimize the end-to-end framework. 
Nevertheless, all of them exhibit suboptimal performance in the presence of artifacts resembling colored and circle dots under extreme bitrates ($\le 0.03$ bpp).
Furthermore, the proposed method achieves bitrate savings ranging from 8.23\% to 11.62\% when compared to HiFiC~\cite{mentzer2020high}, as indicated in Table~\ref{tab:tab1}.
As a second line of approach, the SAVD method~\cite{changIJCV} encodes images into semantic maps and corresponding texture features.
We compare our method with the model trained on the ADE20K outdoor dataset.
It can be observed that it suffers from poor generalization ability in the presence of a semantic gap between the training and testing domains, resulting in inadequate texture generation.
In contrast, our method exhibits superior performance and generalization in generating reconstructed images with both subjective perception and objective metrics, with much lower bitrates.
%

%

\subsection{Lost VQ-indices Prediction}

\label{sec: Lost indices Recovery Task}
Fig.~\ref{fig:mask} presents a visualization of the lost indices map ranging from $10\%$ to $40\%$ loss ratios, as well as the corresponding restored images, generated via the utilization of indices predicted by the transformer model.
It can be observed that the restored images display a similar degree of fidelity to the reconstructed image without missing indices when the levels of missing data are not particularly significant ($\le 20\%$).
Large ratio loss can also preserve the rough structure of the original image, which verifies the robustness and resilience of the proposed coding framework.

\vspace{-4 mm}
\section{Conclusions}
In this work, we propose a novel scheme that utilizes the VQ-indices maps obtained from VQGANs as compact visual data representations to achieve extremely high image compression ratios while maintaining perceptual quality. 
Our proposed scheme uniquely adjusts the quantization step by varying the size of codebooks through the K-means clustering and recovers missing indices by the transformer, enabling reliable compression with variable bit rates and varying levels of reconstruction quality.
Qualitative and quantitative results demonstrate the superiority of our proposed scheme in perceptual quality at extremely low bit rates ($\le 0.04$ bpp) compared to state-of-the-art codecs. 
Even when up to $20\%$ of indices are lost, the images can be successfully restored with minimal perceptual loss.
Overall, our work advances image/video coding research by demonstrating the potential of VQ-based generative models for research in ultra-low bitrate compression.

\Section{References}
\bibliographystyle{IEEEbib}
\bibliography{refs}
\end{document}